\documentclass[twocolumn,aps,showpacs,pra,groupedaddress]{revtex4}
\usepackage{epsfig,amssymb,amsmath}
\usepackage[hypertex,linkcolor=red]{hyperref}
\def\comment#1{}

\begin{document}
\draft
\title{The Mysterious Optimality of Naive Bayes:\\Estimation of the Probability in the System of "Classifiers"}
\author{Oleg Kupervasser} \affiliation{Department of Chemical Physics,The Weizmann
Institute of Science,Rehovot 76100, Israel}

\begin{abstract}
Bayes Classifiers are widely used currently for recognition, identification and knowledge discovery. The fields of application are, for example, image processing, medicine, chemistry (QSAR). However, by mysterious way the Naive Bayes Classifier usually gives a very nice and good presentation of recognition. More complex models of Bayes Classifier cannot improve it considerably. We demonstrate here a very nice and simple proof of the Naïve Bayes Classifier optimality that can explain this interesting fact. The derivation in the current paper is based on
paper of the auther written in 2002.
\end{abstract}
\maketitle
\section{Introduction}
The derivation in the current paper is based onpaper of the auther written in 2002 \cite{KUPER953} .
  Bayes Classifiers are widely used currently for recognition, identification and knowledge discovery. The fields of application are, for example, image processing, medicine, chemistry (QSAR). The special significance such Classifiers have in Medical Diagnostics and Bioinformatics. Very nice examples can be found in paper \cite{Kuper11}. However, these Bayes Classifiers have remarkable property - by mysterious way the Naive Bayes Classifier usually gives a very nice and good presentation of recognition. More complex models of Bayes Classifier \cite {Kuper12} cannot improve it considerably. 

  Let us give some example from practices of author. The first example was recognition of digits written by hand. Every such digit can be characterized by set of variables. The second example is defect on computer screen – scratches, air bubbles, cavities, spots. They can be characterized by set of variables, for example, square of circumscribed ellipse, its eccentricity and so on. The third example is medical diagnostics. We must recognize the diseases on basis of medical symptoms. The all three examples had the same property: in spite of the fact that correlations exist between characteristic variables, the Naive Bayes model gave the excellent result. Moreover, this result could not be improved considerably by using more complex model with some correlations between characteristic variables. Sometimes these correlations (if they are found with errors) can make the model even worse.

In the paper \cite{Kuper12} authors explain this remarkable property. However, they use some assumption (Zero-One Loss) which decreases universality and generality of this consideration. We give in this paper a general proof Naive Bayes Classifier optimality. The derivation in the current paper is similar to \cite{KUPER953} (2002). The subsequent interesting development of the problem was made in \cite{Kuper13} (2004), \cite{Kuper14} (2006). However, unfortunately these papers do not include any analysis of previous one \cite{KUPER953}.

 Let us formulate shortly the basic problem that we try to solve in the paper. Suppose that we have a set of some objects and a set of variables that characterize these objects. For every object, we know probability distribution for every variable. However, we have no information about correlations of the variables. Now, suppose that we know variables values for some sample of the objects. What is probability that this sample correspond to some object? It is a typical problem of recognition over a condition of incomplete information.

 Let us consider the simplest case when no correlations exist between variables. In this case, the Naive Bayes model is an exact solution of the problem. We prove in this paper that for the case that we know nothing about correlation – the Naive Bayes model is not exact, but $optimal$ solution in some sense. More detailed, we prove that the Naive Bayes model gives minimal mean error over all possible models of correlation. We suppose that all correlations models have the equal probability. We think that this result can explain the described above mysterious optimality of Naive Bayes.

  The paper is organized as following. In section II we give exact mathematical definition of the problem for two variables and two objects. In section III we define our notations. In section IV we give generic form of conditional probability for all possible correlations of our variables. In section V we define the restrictions of the functions describing the correlations. In section VI we give the definition a distance between two probability(correlation) models. In section VII we find restrains for our basic functions. In the section VIII we solve our main problem – we prove optimality of the Naive Bayes model for uniform distribution of all possible correlations. In the section IX we find mean error between the Naive Bayes model and an actual model for uniform distribution of all possible correlations. In section X we consider the case more than two variables and objects. The last section is conclusions.

\section{Statement of the problem.} 
Let A be a random variable, with values in set ${0,1}$. Assume that the $a priori$ probability $P(A)=P(A=1)$ is known and denote it by $\theta$. Let $X_1,X_2$ be two random variables, with values in some set, say $]-\infty; +\infty[$. We are given the following information:$X_1=x_1$ and $X_2=x_2$ (obtained though measurement). Furthermore, we have two systems - "classifiers" , which given $x_1$ and $x_2$ produce:

\begin{equation}
P(A=1/X_1=x_1)=P(A/x_1) \doteq \alpha
\end{equation}

\begin{equation}
P(A=1/X_2=x_2)=P(A/x_2) \doteq \beta
\end{equation}

\begin{figure}[htb]
 \psfig{figure=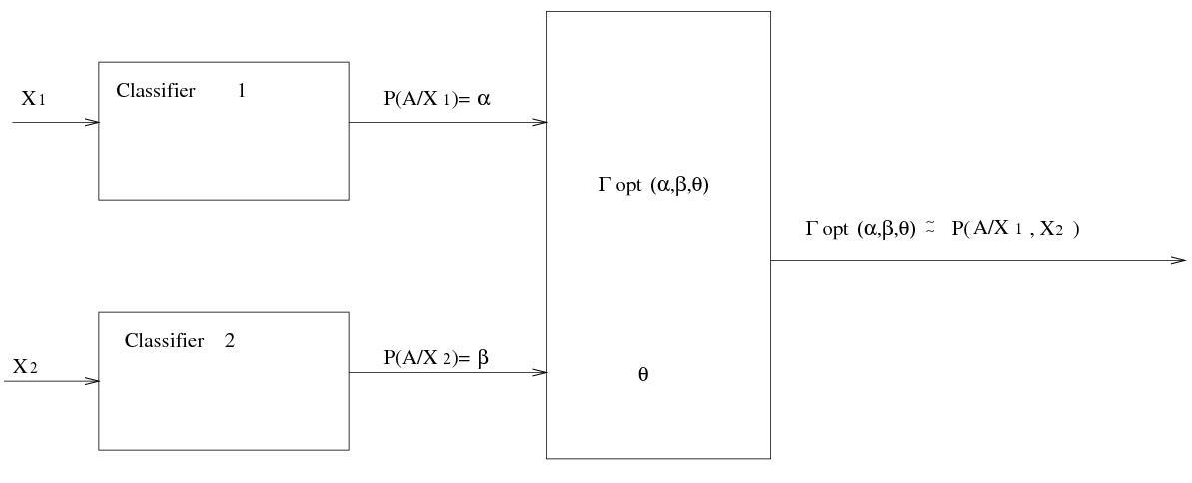, width=6 in}
 \caption {Function $\Gamma(\alpha,\beta,\theta):[0,1]^3 \mapsto
[0,1]$ } \label{file=Fig.1}
\end{figure}

We wish to estimate the probability $P(A=1/X_1=x_1,X_2=x_2)=P(A/x_1,x_2)$ in terms of $\alpha, \beta$ and $\theta$. More specifically we wish to find a function $\Gamma_{opt}(\alpha,\beta,\theta)$  which on the average is the best approximation for $P(A/x_1,x_2)$ in a sense to be defined explicitly in the sequel (see FIG. \ref{file=Fig.1}.).

\section{Notation and preliminaries}

$\rho_{X_1,X_2}(x_1,x_2)$ - joint PDF(probability density function) of
$X_1$ and $X_2$. $\rho_{X_1,X_2/A}(x_1,x_2) \doteq h(x_1,x_2) $ -joint PDF of $X_1$ and $X_2$ , given $A=1$. In terms of $h(x_1,x_2)$ and $\theta$ we may write $P(A/x_1,x_2)$ as follows:

\begin{equation}
P(A/x_1,x_2)={\theta h(x_1,x_2) \over \theta h(x_1,x_2)+(1-\theta)
\overline{h} (x_1,x_2)} \label{1hil}
\end{equation}

where

$\overline{h} (x_1,x_2) \doteq \rho_{X_1,X_2/\overline{A}}(x_1,x_2)$-
joint PDF of $X_1$ and $X_2$, given $A=0$.

We have:
\begin{equation}
\rho_{X_1}(x_1)=\int_{-\infty}^{+\infty}\rho_{X_1,X_2}(x_1,x_2)dx_2
\end{equation}

\begin{equation}
\rho_{X_2}(x_2)=\int_{-\infty}^{+\infty}\rho_{X_1,X_2}(x_1,x_2)dx_1
\end{equation}

\begin{equation}
h_1(x_1) \doteq \rho_{X_1/A}(x_1)=\int_{-\infty}^{+\infty}h(x_1,x_2)dx_2
\end{equation}

\begin{equation}
h_2(x_2) \doteq \rho_{X_2/A}(x_2)=\int_{-\infty}^{+\infty}h(x_1,x_2)dx_1
\end{equation}

\begin{equation}
\overline{h}_1(x_1) \doteq \rho_{X_1/\overline{A}}(x_1)
=\int_{-\infty}^{+\infty}\overline{h}(x_1,x_2)dx_2
\end{equation}

\begin{equation}
\overline{h}_2(x_2) \doteq \rho_{X_2/\overline{A}}(x_2)
=\int_{-\infty}^{+\infty}\overline{h}(x_1,x_2)dx_1
\end{equation}

\section{Generic form of $P(A/x_1,x_2)$}
Define the function $g(x_1,x_2)$ and $\overline{g}((x_1,x_2)$

\begin{equation}
g(x_1,x_2) \doteq {h(x_1,x_2) \over h_1(x_1) h_2(x_2)}
\end{equation}

\begin{equation}
 \overline{g}(x_1,x_2) \doteq {\overline{h}(x_1,x_2) \over
 \overline{h}_1(x_1) \overline{h}_2(x_2)}
\end{equation}

Note that if $X_1$ and $X_2$ are conditionally independent, i.e.
\begin{eqnarray}
 &&h(x_1,x_2)=\rho_{X_1X_2/A}(x_1,x_2)=\rho_{X_1/A}(x_1)\rho_{X_2/A}(x_2)
 \nonumber \\=
 &&h_1(x_1)h_2(x_2)
\end{eqnarray}

then

\begin{equation}
g(x_1,x_2)=\overline{g}(x_1,x_2)=1
\end{equation}

Define the following $monotonously$ $nondecreasing$ probability distribution functions

\begin{equation}
H_1(x_1) \doteq \int_{-\infty}^{x_1} h_1(z)dz
\end{equation}

\begin{equation}
H_2(x_2) \doteq \int_{-\infty}^{x_2} h_2(z)dz
\end{equation}

\begin{equation}
\overline{H}_1(x_1) \doteq \int_{-\infty}^{x_1} \overline{h}_1(z)dz
\end{equation}

\begin{equation}
\overline{H}_2(x_2) \doteq \int_{-\infty}^{x_2} \overline{h}_2(z)dz
\end{equation}

Note that since $H_1(x_1),H_2(x_2),\overline{H}_1(x_1)$ and $ \overline{H}_2(x_2)$ are monotonous (At this point one could assume that $h_1(x_1),h_2(x_2),\overline{h}_1(x_1),\overline{h}_2(x_2) >0$, so that $H_1(x_1),H_2(x_2),\overline{H}_1(x_1)$ and $\overline{H}_2(x_2)$ are monotonously increasing. This restriction will be shown to be superfluous in the sequel.) , there exist the inverse functions $H_1^{-1}(x_1),H_2^{-1}(x_2),
\overline{H}_1^{-1}(x_1)$ and $\overline{H}_2^{-1}(x_2)$. We may therefore define:

\begin{equation}
J(a,b) \doteq g(H_1^{-1}(a),H_2^{-1}(b))
\end{equation}

\begin{equation}
\overline{J}(a,b) \doteq \overline{g}
(\overline{H}_1^{-1}(a),\overline{H}_2^{-1}(b))
\end{equation}

For the sake of brevity we shall henceforth denote

\begin{eqnarray}
&&J \doteq J(H_1(x_1),H_2(x_2))=\nonumber \\
&&g(H_1^{-1}(H_1(x_1)),H_2^{-1}(H_2(x_2)))
=g(x_1,x_2)
\end{eqnarray}

\begin{eqnarray}
&&\overline{J} \doteq \overline{J}(\overline{H}_1(x_1),\overline{H}_2(x_2))=
\nonumber \\
&&\overline{g}(\overline{H}_1^{-1}(\overline{H}_1(x_1)),\overline{H}_2^{-1}
(\overline{H}_2(x_2)))
=\overline{g}(x_1,x_2)
\end{eqnarray}

By the definition

\begin{equation}
h(x_1,x_2)=J h_1(x_1) h_2(x_2)  \label{tito}
\end{equation}

\begin{equation}
\overline{h}(x_1,x_2)=\overline{J} \overline{h}_1(x_1) \overline{h}_2(x_2)
 \label{tito1}
\end{equation}

We now have:

\begin{equation}
h_1(x_1) \doteq \rho_{X_1/A}(x_1)={\rho_{X_1}(x_1)P(A/x_1) \over P(A)}=
 {\alpha \rho_{X_1}(x_1) \over \theta} \ . \label{tito2}
\end{equation}

\begin{equation}
h_2(x_2) \doteq \rho_{X_2/A}(x_2)={\rho_{X_2}(x_2)P(A/x_2) \over P(A)}=
 {\beta \rho_{X_2}(x_2) \over \theta} \ . \label{tito3}
 \end{equation}

\begin{eqnarray}
&&\overline{h}_1(x_1) \doteq \rho_{X_1/\overline{A}}(x_1)={\rho_{X_1}(x_1)
P(\overline{A}/x_1)
\over P(\overline{A})}=
\nonumber \\
&&{(1-\alpha) \rho_{X_1}(x_1) \over 1-\theta}  \label{tito4}
\end{eqnarray}

\begin{eqnarray}
&&\overline{h}_2(x_2) \doteq \rho_{X_2/\overline{A}}(x_2)={\rho_{X_2}(x_2)
P(\overline{A}/x_2)
\over P(\overline{A})}=
\nonumber \\
&&{(1-\alpha) \rho_{X_2}(x_2) \over 1-\theta}  \label{tito5}
\end{eqnarray}

Hence, from (\ref{tito}),(\ref{tito1})

\begin{equation}
h(x_1,x_2)=J{\alpha \beta \rho_{X_1}(x_1) \rho_{X_2}(x_2) \over \theta^2}
\end{equation}

\begin{equation}
\overline{h}(x_1,x_2)=\overline{J}{(1-\alpha) (1-\beta)
\rho_{X_1}(x_1) \rho_{X_2}(x_2) \over (1-\theta)^2}
\end{equation}

Now from (\ref{1hil})

\begin{eqnarray}
&&P(A/x_1,x_2)=\nonumber \\
&&{ {J \over \theta} \alpha \beta \rho_{X_1}(x_1) \rho_{X_2}(x_2)
\over {J \over \theta}\alpha \beta \rho_{X_1}(x_1) \rho_{X_2}(x_2)+
{\overline{J}\over (1-\theta)}(1-\alpha) (1-\beta)
\rho_{X_1}(x_1) \rho_{X_2}(x_2)}
\nonumber \\ &&
={\alpha \beta \over \alpha \beta+{\overline{J} \over J} {\theta \over
1- \theta} (1-\alpha)(1- \beta)} \label{kmax}
\end{eqnarray}

Note that in case of conditional independence $J=\overline{J}=1$ and
(\ref{kmax}) becomes the exact solution $\Gamma(\alpha,\beta,\theta)=
P(A/x_1,x_2)$.

\section{Restrictions on the functions $J(a,b)$ and $\overline{J}(a,b)$}

We have

\begin{equation}
h_1(x_1)=\int^{+\infty}_{-\infty}J(H_1(x_1),H_2(x_2))h_1(x_1)h_2(x_2)dx_2
\ . \label{kmax1}
\end{equation}

Hence

\begin{eqnarray}
&&1=\int^{+\infty}_{-\infty}J(H_1(x_1),H_2(x_2))h_2(x_2)dx_2=
\nonumber \\
&&\int^{1}_{0}J(H_1(x_1),H_2(x_2))dH_2(x_2)  \label{kmax2}
\end{eqnarray}

Thus, we have the following condition

\begin{equation}
\int^{1}_{0}J(a,b)db=1  \label{kmax3}
\end{equation}

and analogously

\begin{equation}
\int^{1}_{0}J(a,b)da=1  \label{kmax4}
\end{equation}

Similarly, we obtain:

\begin{eqnarray}
&&\int^{1}_{0}\overline{J}(a,b)da=1\nonumber \\ &&
\int^{1}_{0}\overline{J}(a,b)db=1  \label{kmax5}
\end{eqnarray}

Obviously

\begin{equation}
J(a,b),\overline{J}(a,b) \geq 0 \ . \label{kmax6}
\end{equation}

\begin{equation}
\int^{1}_{0}\int^{1}_{0}J(a,b)dadb=
\int^{1}_{0}\int^{1}_{0}\overline{J}(a,b)dadb=1 \ . \label{kmax7}
\end{equation}

The set of all the solutions of (\ref{kmax3}),(\ref{kmax4}),(\ref{kmax5}),(\ref{kmax6}),(\ref{kmax7}) together with (\ref{kmax}) determines the set of all possible realizations of $P(A/x_1,x_2)$.

 $An$ $example$ $of$ $a$ $solution$ $of$ (\ref{kmax3}),(\ref{kmax4}) $and$
 (\ref{kmax6}),(\ref{kmax7}).

 Let $\rho(x)$ be a function such that $\rho(x) \geq 0$ and
 $\int_0^1\rho(x)dx=1$

 Then

\begin{equation}
J(a,b)= \left \{ \begin{array}{cc}
 \rho(a-b)&,a \geq b \\\rho(a-b+1)&,a<b\end{array}\right.
\end{equation}

satisfies (\ref{kmax3}),(\ref{kmax4}) and (\ref{kmax6}),(\ref{kmax7}).

\section{Definition of distance}

We define the distance between the proposed approximation of $P(A/x_1,x_2)$,- $\Gamma(\alpha,\beta,\theta)$ and the actual function $P(A/x_1,x_2)$ as follows:

\begin{eqnarray}
&&||\Gamma(\alpha,\beta,\theta)-P(A/x_1,x_2)|| \doteq \nonumber \\
&&{\int \int}{-\infty}
^{+\infty} \rho_{X_1X_2}(x_1,x_2) \nonumber \\&&
[\Gamma(\alpha,\beta,\theta)-P(A/x_1,x_2)]^2
dx_1dx_2
\end{eqnarray}

Now we have from (\ref{tito}),(\ref{tito1}) and (\ref{tito2}),(\ref{tito3}),
(\ref{tito4}),(\ref{tito5})

\begin{eqnarray}
&&\rho_{X_1X_2}(x_1,x_2)=\theta h(x_1,x_2)+(1-\theta)\overline{h}(x_1,x_2)=
\nonumber \\
&&\theta J h_1(x_1) h_2(x_2)+ (1-\theta)\overline{J} \overline{h}_1(x_1)
\overline{h}_2(x_2)= \nonumber \\
&&[{J \alpha \beta  \over \theta}+{\overline{J}(1-\alpha)
(1-\beta) \over (1-\theta)}] \rho_{X_1}(x_1) \rho_{X_2}(x_2)
\end{eqnarray}

\begin{eqnarray}
&&||\Gamma(\alpha,\beta,\theta)-P(A/x_1,x_2)||\nonumber \\ &&
={\int \int}_{-\infty}^{+\infty}
\rho_{X_1}(x_1) \rho_{X_2}(x_2) \nonumber \\ &&
[{J \alpha \beta  \over \theta}+
{\overline{J}(1-\alpha)(1-\beta) \over (1-\theta)}]\nonumber \\ &&
(\Gamma(\alpha,\beta,\theta)-P(A/x_1,x_2))^2dx_1dx_2
\nonumber \\ &&=
{\int}_{0}^1 {\int}_{0}^1[{J \alpha \beta  \over \theta}+
{\overline{J}(1-\alpha)(1-\beta) \over (1-\theta)}]\nonumber \\ &&
(\Gamma(\alpha,\beta,\theta)-P(A/x_1,x_2))^2dF_1(x_1)dF_2(x_2)
 \label{popka}
\end{eqnarray}

where

\begin{equation}
F_1(x_1)=\int^{x_1}_{-\infty}\rho_{X_1}(z)dz
\end{equation}

\begin{equation}
F_2(x_2)=\int^{x_2}_{-\infty}\rho_{X_2}(z)dz
\end{equation}

\section{ Restraints for basic functions}

We will consider in further all functions with arguments $1 \geq F1,F2 \geq 0$, but not $x_1,x_2$. We have six function of $F1,F2$, that define (\ref{popka}): $J,\overline{J},H_1,H_2,\alpha,\beta$. Let us to write the other function by help these function and find restraints for these functions.

(i)
\begin{equation}
\alpha=P(A/x_1)=\theta h_1(x_1)/\rho_{X_1}(x_1)=\theta {{dH_1 \over dx_1}
\over {dF_1 \over dx_1}}=\theta {dH_1 \over dF_1}
\end{equation}

By the same way

\begin{equation}
\beta=\theta {dH_2 \over dF_2}
\end{equation}

We know that functions $H_1, F_1, H_2, F_2$ are cumulative distribution functions of  $x_1$,$x_2$, correspondently.
These functions are $monotonously$ $nondecreasing$ functions and changes from 0 to 1 from the definition of cumulative distribution functions.
Therefore, we can conclude the following restraints for functions $H_1,H_2$ as functions of $F_1,F_2$ exist :

\begin{eqnarray}
&&H_1(1)=H_2(1)=1 \nonumber\\ &&
H_1(0)=H_2(0)=0
\end{eqnarray}

\begin{equation}
0 \leq \alpha=\theta {dH_1 \over dF_1},\beta=\theta {dH_2 \over dF_2} \leq 1
\end{equation}

\begin{equation}
0 \leq \theta \leq 1
\end{equation}

(ii)

\begin{eqnarray}
&& \overline{H}_1(x_1)=\int_{-\infty}^{x_1}\overline{h}_1(x_1)=
\int_{-\infty}^{x_1}{(1-\alpha) \rho_{X_1}(x_1) \over 1-\theta}dx_1=
\nonumber \\ &&
{1 \over 1-\theta}\int_{-\infty}^{x_1}- {\theta \over 1-\theta}
\int_{-\infty}^{x_1} { \alpha \rho_{X_1}(x_1) \over \theta}dx_1
\nonumber \\ && =
{F_1 \over 1-\theta}- {\theta \over 1-\theta}H_1(x_1)
\end{eqnarray}

By the same way

\begin{equation}
\overline{H}_2(x_2)=
{F_2 \over 1-\theta}- {\theta \over 1-\theta}H_2(x_2)
\end{equation}

(iii)

\begin{eqnarray}
&&J(H_1(F_1),H_2(F_2)): \nonumber \\&&
J(H_1(F_1),H_2(F_2)) \geq 0 \nonumber \\&&
\int^{1}_{0}J(a,b)db=1 \nonumber \\ &&
\int^{1}_{0}J(a,b)da=1
\end{eqnarray}

\begin{eqnarray}
&&\overline{J}(\overline{H}_1(F_1),\overline{H}_2(F_2)):\nonumber \\&&
\overline{J}(\overline{H}_1(F_1),\overline{H}_2(F_2)) \geq 0 \nonumber \\&&
\int^{1}_{0}\overline{J}(a,b)db=1 \nonumber \\&&
\int^{1}_{0}\overline{J}(a,b)da=1
\end{eqnarray}

(iv)

\begin{equation}
P(A/x_1,x_2)=
{{J\alpha \beta \over \theta}
\over {J\alpha \beta \over \theta}+ {\overline{J} (1-\alpha)(1- \beta) \over
1- \theta }} \label{eqz1}
\end{equation}

\section{Optimization}

We shell find the best approximation $\Gamma(\alpha,\beta,\theta)$ as follows

\begin{equation}
min_{\Gamma(\alpha,\beta,\theta)} E[||\Gamma(\alpha,\beta,\theta)-
P(A/x_1,x_2)||] \longrightarrow \Gamma(\alpha,\beta,\theta)
\end{equation}

where the expected value (or expectation, or mathematical expectation, or mean, or the first moment)  $E[...]$  is taken with respect to the joint PDF of possible realizations of: $J, \overline{J}, \alpha, \beta, H_1, H_2$ for given $F_1$ and $F_2$.

For the sake of brevity, we denote:

\begin{equation}
C \doteq {J \alpha \beta \over \theta}+
{\overline{J} (1-\alpha) (1-\beta) \over (1-\theta)}
\end{equation}

\begin{equation}
D \doteq {J \alpha \beta \over \theta}
\end{equation}

Then from(\ref{eqz1}) and (\ref{popka})

\begin{eqnarray}
&&||\Gamma(\alpha,\beta,\theta)-P(A/x_1,x_2)||=\nonumber \\&&
\int_0^1 \int_0^1
C(\Gamma(\alpha,\beta,\theta)-D/C)^2dF_1dF_2=\nonumber \\&&
\int_0^1 \int_0^1 dF_1dF_2 [{D^2 \over C}
+ \Gamma^2(\alpha,\beta,\theta)C
-2 \Gamma(\alpha,\beta,\theta) D]\label{eqz6}
\end{eqnarray}

Thus

\begin{eqnarray}
&&min_{\Gamma(\alpha,\beta,\theta)} E[||\Gamma(\alpha,\beta,\theta)-
P(A/x_1,x_2)||]= \nonumber \\&&
min_{\Gamma(\alpha,\beta,\theta)} E[\int_0^1 \int_0^1 dF_1dF_2\nonumber \\&&[
{D^2 \over C} + \Gamma^2(\alpha,\beta,\theta)C
-2 \Gamma(\alpha,\beta,\theta) D]]= \nonumber \\&&
min_{\Gamma(\alpha,\beta,\theta)} E[\int_0^1 \int_0^1 dF_1dF_2[{D^2 \over C}]]
+\nonumber \\&&
min_{\Gamma(\alpha,\beta,\theta)} E[\int_0^1 \int_0^1 dF_1dF_2\nonumber \\&&[
 \Gamma^2(\alpha,\beta,\theta)C
-2 \Gamma(\alpha,\beta,\theta) D]]= \nonumber \\&&
Const+
min_{\Gamma(\alpha,\beta,\theta)} E[\int_0^1 \int_0^1 dF_1dF_2\nonumber \\&&[
 \Gamma^2(\alpha,\beta,\theta)C
 -2 \Gamma(\alpha,\beta,\theta) D]]\label{eqz2}
\end{eqnarray}

It remains to calculate the expected value in (\ref{eqz2}).

We have by obvious assumptions

\begin{eqnarray}
&&\rho_{J,\overline{J},\alpha,\beta,H_1,H_2/F_1,F_2}
(J,\overline{J},\alpha,\beta,H_1,H_2/F_1,F_2)= \nonumber \\&&
\rho_{J/H_1,H_2}(J/H_1,H_2) \rho_{\overline{J}/\overline{H}_1,\overline{H}_2}
(\overline{J}/\overline{H}_1,\overline{H}_2) \nonumber \\&&
\rho_{\alpha/F_1}(\alpha/F_1) \rho_{H_1/\alpha,F_1}(H_1/\alpha,F_1)
\nonumber \\&&
\rho_{\beta/F_2}(\beta/F_2) \rho_{H_2/\beta,F_2}(H_2/\beta,F_2)
\label{eqz4}
\end{eqnarray}

\subsection{Lemma 1}

\begin{equation}
E[J(a,b)]=\int_0^{+\infty} \rho_{J(a,b)/a,b}(J(a,b)/a,b) J(a,b) dJ=1
\end{equation}

\begin{equation}
E[\overline{J}(a,b)]=
\int_0^{+\infty} \rho_{\overline{J}(a,b)/a,b}(\overline{J}(a,b)/a,b)
\overline{J}(a,b) d\overline{J}=1
\end{equation}

Proof:

Let us consider function: $\rho_{J(a,b)/a,b}$. Function $J(a,b)$ is defined on the square $0 \leq a,b \leq 1$. Let us make sampling of function $J$ on this square by its dividing on small squares $(i,j)$ and define value of the function $J_{ij}$ on every square $i,j$. Restraints for function $J$ (***) can be written

\begin{equation}
J_{ij} \geq 0
\end{equation}

\begin{equation}
{1 \over N}\sum_{i=1}^N J_{ij} =1
\end{equation}

\begin{equation}
{1 \over N}\sum_{j=1}^N J_{ij} =1
\end{equation}

here $i=1,...,N$, $j=1,...,N$

All matrixes $(J_{ij})$ that satisfy these conditions  are equal probability.
Let us define probability density function

\begin{equation}
\rho(J_{11},...,J_{ij},...,J_{NN})
\end{equation}

This density function must be symmetric with respect to transpositions lines and columns in matrix $(J_{ij})$ , because the density function has equal probability for all matrixes $(J_{ij})$ that satisfy the above conditions.
Indeed, these conditions are also symmetric with respect to transpositions lines and columns in matrix $(J_{ij})$. From symmetry conditions that define this function $(\rho)$ with respect to transpositions lines and columns in matrix $(J_{ij})$ we can conclude that this function $(\rho)$ also doesn't transform with respect to such transpositions.

Let us consider function $\rho_{u/ij}(u/ij)$ which is a discrete version
of the function $\rho_{J(a,b)/a,b}(J(a,b)/a,b)$ :

\begin{eqnarray}
&&\rho_{u/ij}(u/ij)={\int ... \int}_{0}^{+\infty}\rho(J_{11},...,J_{nk},
...,J_{ij}=u,...,J_{NN}) \nonumber \\&&
\prod_{(lm) \neq (ij)} dJ_{lm}
\end{eqnarray}

Let us transpose lines and columns $ (J_{ij})$ by such way that element $J_{ij}$ will be replaced by element $J_{nk}$ , the function $\rho(J_{11},...)$ will not be transform after it. So from previous equation we obtain

\begin{eqnarray}
&&\rho_{u/ij}(u/ij)={\int ... \int}_{0}^{+\infty}\rho(J_{11},...,J_{nk}=u,
...,J_{ij},...,J_{NN}) \nonumber \\&&
\prod_{(lm) \neq (nk)} dJ_{lm}=\rho_{u/nk}(u/nk)
\end{eqnarray}

From this equation we can conclude that $\rho_{u/ij}(u/ij)$ doesn't depend on $ij$ so $\rho_{J/ab}(J/ab)$ doesn't depend on $ab$ and

\begin{equation}
\rho_{J/ab}(J/ab)=\rho_{J}(J)
\end{equation}

and

\begin{equation}
E[J(a,b)]=\int_0^{+\infty} \rho_{J}(J) J dJ=Const \label{eqz3}
\end{equation}

from

\begin{equation}
\int_0^1 \int_0^1 J(a,b)da db=1
\end{equation}

we can conclude that

\begin{equation}
\int_0^1 \int_0^1 E[J(a,b)]dadb=1
\end{equation}

So we can obtain that $Const=1$ in Eq.(\ref{eqz3}).

\subsection{Lemma 2}

Probability distribution functions $\alpha$ and $\beta$ do not dependent on $F_1$ and $F_2$.

\begin{equation}
\rho_{\alpha/F_1}(\alpha/F_1)=\rho_{\alpha}(\alpha)
\end{equation}

\begin{equation}
\rho_{\beta/F_2}(\beta/F_2)=\rho_{\beta}(\beta)
\end{equation}

Proof:

Let us make sampling of function $\alpha(F_1)$ by dividing of domain of this function $F_1,[0,1]$ on intervals of $1/N,N \gg 1$. Then restriction conditions for $\alpha_k,k=1,...,N$:

\begin{equation}
0 \leq \alpha_k \leq 1
\end{equation}

\begin{equation}
 {1 \over N} \sum_{k=1}^N\alpha_k=\int_0^1 \theta {d H_1(F_1) \over d F_1}
 d F_1=\theta
\end{equation}

All columns ($\alpha_k$) that satisfy by this conditions are equal probability.
Let us to consider respective function $\rho(\alpha_1,...,\alpha_k,...,\alpha_l,...,\alpha_N)$. From symmetry conditions that define this function with respect to transpositions
 $\alpha_k \rightarrow \alpha_l$ function
 $\rho(\alpha_1,...,\alpha_k,...,\alpha_l,...,\alpha_N)$ also doesn't
 transform with respect to such transpositions. So we can write

 \begin{eqnarray}
 &&\rho_k(u)=\int_0^1 \rho(\alpha_1,...,\alpha_k=u,...,\alpha_l,...,\alpha_N)
 \prod_{n \neq k}d\alpha_n=\nonumber \\&&
 \int_0^1 \rho(\alpha_1,...,\alpha_k,...,\alpha_l
 =u,...,\alpha_N)\prod_{n \neq l}d\alpha_n=
 \nonumber \\&& \rho_l(u)
\end{eqnarray}

 From this equation, we can conclude that function $\rho_{\alpha/F_1}(\alpha/F_1)$ doesn't depend on $F_1$.

\begin{equation}
\rho_{\alpha/F_1}(\alpha/F_1)=\rho_{\alpha}(\alpha)
\end{equation}

From (\ref{eqz4}) we obtain

\begin{eqnarray}
&&E[\Gamma^2(\alpha,\beta,\theta)C-2\Gamma(\alpha,\beta,\theta)D]=
\nonumber \\&&
\int_0^1\int_0^1\rho_{\alpha}(\alpha)\rho_{\beta}(\beta)d\alpha d\beta
\nonumber \\&&
\int_0^1\int_0^1\rho_{H_1/\alpha,F_1}(H_1/\alpha,F_1)
\rho_{H_2/\beta,F_2}(H_2/\beta,F_2)dH_1 dH_2\nonumber \\&&
\int_0^{\infty} \int_0^{\infty} \rho_{J}(J) \rho_{\overline{J}}(\overline{J})
[\Gamma^2(\alpha,\beta,\theta)[{J\alpha \beta \over \theta}+{
\overline{J}(1-\alpha)(1-\beta) \over 1-\theta}]
\nonumber \\&&
-2\Gamma(\alpha,\beta,\theta)
{J \alpha \beta \over \theta}]dJ d\overline{J}
\nonumber \\&&
=\int_0^1\int_0^1\rho_{\alpha}(\alpha)\rho_{\beta}(\beta)d\alpha d\beta
\nonumber \\&&
[\Gamma^2(\alpha,\beta,\theta)[{E[J]\alpha \beta \over \theta}+{
E[\overline{J}](1-\alpha)(1-\beta) \over 1-\theta}]
\nonumber \\&&
-2\Gamma(\alpha,\beta,\theta)
{E[J] \alpha \beta \over \theta}]
\end{eqnarray}

Let us define

\begin{equation}
\overline{C}={\alpha \beta \over \theta} +{(1-\alpha) (1-\beta) \over 1-\theta}
\end{equation}

\begin{equation}
\overline{D}={\alpha \beta \over \theta}
\end{equation}

By Lemma 1, $E[J]=E[\overline{J}]=1$. Hence

\begin{eqnarray}
&&E[\Gamma^2(\alpha,\beta,\theta)C-2\Gamma(\alpha,\beta,\theta)D]=
\int_0^1\int_0^1
[\Gamma^2(\alpha,\beta,\theta)\overline{C}
\nonumber \\&&
-2\Gamma(\alpha,\beta,\theta)\overline{D}]
\rho_{\alpha}(\alpha)\rho_{\beta}(\beta)d\alpha d\beta
\end{eqnarray}

It remain to find

\begin{eqnarray}
&&min_{\Gamma(\alpha,\beta,\theta)} \int_0^1\int_0^1 dF_1 dF_2
\nonumber \\&&
\int_0^1\int_0^1 d\alpha d\beta \rho_{\alpha}(\alpha) \rho_{\beta}(\beta)
\nonumber \\&&
[\Gamma^2(\alpha,\beta,\theta)\overline{C}
-2\Gamma(\alpha,\beta,\theta)\overline{D}] \label{eqz5}
\end{eqnarray}

Since

\begin{equation}
\rho_{\alpha}(\alpha) \rho_{\beta}(\beta) \geq 0
\end{equation}

if the expression in square brackets is minimized at each point then the whole integral in (\ref{eqz5}) is minimized. Thus, we may proceed as follows

\begin{equation}
{\partial \over \partial \Gamma}
[\Gamma^2(\alpha,\beta,\theta)\overline{C}
-2\Gamma(\alpha,\beta,\theta)\overline{D}]
=2\Gamma(\alpha,\beta,\theta)\overline{C}-2\overline{D}=0
\end{equation}

Hence the optimum $\Gamma(\alpha,\beta,\theta)$ is given by

\begin{equation}
\Gamma_{opt}(\alpha,\beta,\theta)={\overline{D} \over \overline{C}}=
{{\alpha \beta \over \theta} \over
{\alpha \beta \over \theta} +{(1-\alpha) (1-\beta) \over 1-\theta}}
\end{equation}

\section{Mean distance beetwen the proposed approximation of
 $P(A/x_1,x_2)$ ,- $\Gamma(\alpha,\beta,\theta)$ and the actual
 function $P(A/x_1,x_2)$}

 The mean distance from (\ref{eqz6}) is

\begin{eqnarray}
 &&DIS=E[||\Gamma(\alpha,\beta,\theta)-P(A/x_1,x_2)||]=
 \nonumber \\&&
 \int_0^1\int_0^1
 \rho_{\alpha}(\alpha)\rho_{\beta}(\beta)d\alpha d\beta
 \nonumber \\&&
  [\Gamma^2(\alpha,\beta,\theta)\overline{C}
   -2\Gamma(\alpha,\beta,\theta)\overline{D}]
   \nonumber \\&&
   +Const
 \end{eqnarray}

 where $Const$ in this equation is defined by

 \begin{eqnarray}
 &&Const=
 \nonumber \\&&
 E[\int_{-\infty}^{+\infty} \int_{-\infty}^{+\infty}
 \rho_{X_1,X_2}(x_1,x_2)[P(A/x_1,x_2)]^2 dx_1 dx_2]
 \end{eqnarray}

 From this equation we can find boundaries of the $Const$.
 From $0 \leq P(A/x_1,x_2) \leq 1$ we can conclude

 \begin{eqnarray}
 &&Const \leq \nonumber \\&&
 E[\int_{-\infty}^{+\infty} \int_{-\infty}^{+\infty}
  \rho_{X_1,X_2}(x_1,x_2)P(A/x_1,x_2) dx_1 dx_2]\nonumber \\&&=
  E[\theta]=\theta
  \end{eqnarray}

  The second condition is

  \begin{eqnarray}
 && 0 \leq E[\int_{-\infty}^{+\infty} \int_{-\infty}^{+\infty}
   \rho_{X_1,X_2}(x_1,x_2)\nonumber \\&&
   [P(A/x_1,x_2)-\theta]^2 dx_1 dx_2]=
 \nonumber \\&&
   E[\int_{-\infty}^{+\infty} \int_{-\infty}^{+\infty}
      \rho_{X_1,X_2}(x_1,x_2)
 \nonumber \\&&
      [P(A/x_1,x_2)^2+\theta^2-2P(A/x_1,x_2)\theta]
      dx_1 dx_2]=
 \nonumber \\&&
      E[\int_{-\infty}^{+\infty} \int_{-\infty}^{+\infty}
     \rho_{X_1,X_2}(x_1,x_2)[P(A/x_1,x_2)]^2 dx_1 dx_2]
\nonumber \\&&
     -\theta^2
  \end{eqnarray}

 So from these two equations we can conclude

 \begin{equation}
 \theta^2 \leq Const \leq \theta
 \end{equation}

 By next step we would like find function $\rho_{\alpha}(\alpha)$
 ($\rho_{\beta}(\beta)$) in equation for $DIS$.

 Restrictions for function $\alpha(F_1),0 \leq F_1 \leq 1$ are next:

 (i)

 \begin{equation}
 \int_0^1 \alpha(F_1)dF_1=\theta
 \end{equation}

 (ii)

 \begin{equation}
 0 \leq \alpha(F_1) \leq 1
 \end{equation}

 In discrete form (for $ N \rightarrow \infty$) we can rewrite
 $\alpha_{set}=\{\alpha_1,\alpha_2,...,\alpha_N\}$

  (i)

   \begin{equation}
   {1 \over N} \sum_{i=1}^N\alpha_i=\theta
  \end{equation}

   (ii)

    \begin{equation}
     0 \leq \alpha_i \leq 1, i=1,2,...,N
      \end{equation}

    Let us define function $U(\alpha_{set})$ by next way

    \begin{equation}
    U(\alpha_{set})=  \left \{ \begin{array} {ccc}
    \sum_{i=1}^N \alpha_i & for 0 \leq \alpha_i \leq 1 &
    , i=1,2,...,N\\
    +\infty & otherwise\end{array}\right.
    \end{equation}

    \begin{equation}
    U(\alpha_{set})=\sum_{i=1}^N U_i(\alpha_i)
    \end{equation}

    \begin{equation}
    U_i(\alpha_{i})=\left \{\begin{array}{cc} \alpha_i
    &for 0 \leq \alpha_i \leq 1\\
    +\infty& otherwise\end{array}\right.
    \end{equation}

    Then function that satisfies equal probability distribution with considering restrictions (i),(ii) is

    \begin{equation}
    \rho_{\alpha_{set}}(\alpha_{set})={1 \over C}\delta(U(\alpha_{set})-
    N\theta) \label{jok}
    \end{equation}

    where $\delta$ - delta-function of Dirac.

    Constant $C$ define by

    \begin{equation}
    \int_{-\infty}^{+\infty} ... \int_{-\infty}^{+\infty}\rho_{\alpha_{set}}
    (\alpha_{set})d\alpha_1...d\alpha_N=1
    \end{equation}

    It can be proved (see each course of "Statistical mechanics"; transform from microcanonical to canonical distribution) that for $N \mapsto \infty$ distribution (\ref{jok}) is equal to next distribution:

    \begin{equation}
     \rho_{\alpha_{set}}(\alpha_{set})={1 \over Z}e^{-KU(\alpha_{set})}
    \end{equation}

where $Z$ and $K$ can be found from equations

    \begin{equation}
    \int_{-\infty}^{+\infty}...\int_{-\infty}^{+\infty}
    \rho_{\alpha_{set}}(\alpha_{set})d\alpha_1...d\alpha_N=1\label{por1}
    \end{equation}

    \begin{equation}
    \int_{-\infty}^{+\infty}...\int_{-\infty}^{+\infty}U(\alpha_{set})
    \rho_{\alpha_{set}}(\alpha_{set})d\alpha_1...d\alpha_N=N\theta
    \label{por2}
    \end{equation}

Quest function $ \rho_{\alpha}(\alpha)$ can be find by

    \begin{eqnarray}
    &&\rho_{\alpha}(\alpha)=
    \int_{-\infty}^{+\infty}...\int_{-\infty}^{+\infty}
    \rho_{\alpha_{set}}(\alpha_1,...,\alpha_j=\alpha,...,\alpha_N)
    \nonumber \\&&
    \prod_{i=1,i \neq j}^N d\alpha_i=
    \nonumber \\&&
    {1 \over D}e^{-KU_j(\alpha_J=\alpha)}\label{por3}
    \end{eqnarray}

    where

    \begin{equation}
    D^N=Z \label{por4}
    \end{equation}

    From Eqs.(\ref{por1}),(\ref{por2}) we can find

    \begin{equation}
    {1 \over Z}=({K \over 1-e^{-K}})^N \label{por5}
    \end{equation}

    \begin{equation}
    \theta=\Lambda(K) \label{por6}
    \end{equation}

    where $\Lambda(K)$ is decreasing function

    \begin{equation}
    \Lambda(K)=\left \{ \begin{array}{cc}
    1&for K=-\infty\\ 0&
    for K=+\infty\\ 1/2&for K=0
    \\ {1 \over K}-{1 \over e^K-1}&
    otherwise\end{array}\right.
    \end{equation}

    If $K$ is root of Eq (ref{por6}) we can write from Eqs.(\ref{por3}),(\ref{por4}),(\ref{por5}),(\ref{por6}) for function $\rho_{\alpha}(\alpha)$:

       \begin{equation}
       \rho_{\alpha}(\alpha)=\left\{ \begin{array}{c}
       \left\{ \begin{array}{cc}
       For& K=0\\
       1& for 0 \leq \alpha \leq 1\\
       0 &    otherwise
       \end{array}\right.\\
       \left\{ \begin{array}{cc}
           For& K=+\infty\\
       2\delta(\alpha)& 0 \leq \alpha \leq 1\\
       0              & otherwise
       \end{array}\right.\\
       \left\{ \begin{array}{cc}
           For& K=-\infty\\
       2\delta(\alpha-1)& 0 \leq \alpha \leq 1\\
       0               &  otherwise
       \end{array}\right.\\
       \left\{ \begin{array}{cc}
       For& otherwise K\\
       {1 \over D} e^{-K\alpha}& 0 \leq \alpha \leq 1\\
       0                   &    otherwise
       \end{array}\right.
       \end{array}\right.
       \end{equation}

       where $2\int_0^1\delta(\alpha-1)=2\int_0^1\delta(\alpha)=1$
       and

       \begin{equation}
       {1 \over D}={K \over 1-e^{-K}}
       \end{equation}

       \section{The case of more than two variables A and X}

Let A be a random variable, with values in set ${0,1,...,L}$.
Assume that the $apriori$ probability $P(A=i)$ is known and denote it by $\theta_i$, here $i=1,...,L$. Let $X_1,...,X_K$ be two random variables, with values in some set, say $]-\infty; +\infty[$. We are given the following information:$X_1=x_1$,...,$X_K=x_K$ (obtained though measurement). Furthermore, we have two systems - "classifiers", which given $x_1$ ,...,$x_K$ produce:

\begin{equation}
P(A=i/X_j=x_j) \doteq \alpha_{ij}
\end{equation}

We wish to estimate the probability $P(A=i/X_1=x_1,...,X_K=x_K)$ in terms of $\alpha_{ij}$ and $\theta_i$. More specifically we wish to find a function $\Gamma_{opt,M}(\alpha_{ij},\theta_{i})$ which on the average is the best approximation for $P(A=M/x_1,...,x_K)$. By the same way, that in case of two variables we can find that the $\Gamma_{opt,M}(\alpha_{ij},\theta_{i})$ defined by equation

\begin{equation}
\Gamma_{opt,M}(\alpha_{ij},\theta_{i})={(\prod_{j=1}^K \alpha_{Mj})
/\theta_M^{K-1} \over \sum_{i=1}^L (\prod_{j=1}^K \alpha_{ij})/\theta_i^{K-1}}
\end{equation}

We have evidential restraints for $\alpha_{ij}$,$\theta_i$

\begin{eqnarray}
&&0 \leq \alpha_{ij} \leq 1 \nonumber \\&&
\sum_{i=1}^L\alpha_{ij}=1
\end{eqnarray}

\begin{eqnarray}
&&0 \leq \theta_i \leq 1  \nonumber \\&&
\sum_{i=1}^L\theta_i=1
\end{eqnarray}

\section{Conclusions}

We proved successfully that the Naive Bayes model gives minimal mean error over uniform distribution of all possible correlation between characteristic variables. This result can explain the described above mysterious optimality of Naive Bayes. We also found mean error that the Naive Bayes model gives for uniform distribution of all possible correlation.

\noindent {\bf Acknowledgments} We would like to thank Alexsander Vardy and Romanov Alexey Nikolaevich for their help in creating this paper. We also would like to thank anonymous referee for very useful and clear remarks.



{\Large \bf Figures Legends}\\~\\

\end{document}